\documentclass{article}

\usepackage[preprint]{neurips_2025}

\usepackage[utf8]{inputenc}
\usepackage[T1]{fontenc}
\usepackage{hyperref}
\usepackage{url}
\usepackage{booktabs}
\usepackage{amsfonts}
\usepackage{amsmath}
\usepackage{amssymb}
\usepackage{nicefrac}
\usepackage{microtype}
\usepackage{graphicx}
\usepackage{xcolor}

\usepackage{makecell}
\usepackage{listings}
\usepackage{algorithm}
\usepackage{algorithmic}
\usepackage{subcaption}
\usepackage[most]{tcolorbox}
\usepackage{enumitem}

\definecolor{codeblue}{rgb}{0.25,0.5,0.75}
\definecolor{codegray}{rgb}{0.5,0.5,0.5}
\definecolor{codegreen}{rgb}{0.0,0.5,0.0}

\title{Explicit Abstention Knobs for Predictable Reliability in Video Question Answering}

\author{
  Jorge Ortiz \\
  Department of Electrical and Computer Engineering\\
  Rutgers University\\
  New Brunswick, NJ 08901 \\
  \texttt{jorge.ortiz@rutgers.edu}
}

\begin{document}

\maketitle

\begin{abstract}
High-stakes deployment of vision-language models (VLMs) requires selective prediction, where systems abstain when uncertain rather than risk costly errors. We investigate whether confidence-based abstention provides reliable control over error rates in video question answering, and whether that control remains robust under distribution shift. Using NExT-QA and Gemini 2.0 Flash, we establish two findings. First, confidence thresholding provides mechanistic control in-distribution. Sweeping threshold $\varepsilon$ produces smooth risk-coverage tradeoffs, reducing error rates from 23.6\% to 9.4\% at 63.7\% coverage with well-calibrated predictions (ECE = 0.018). Second, this control is not epistemic. Under evidence degradation (18 frames reduced to 6), the model's confidence distribution contracts only modestly. Evaluating the same frozen question instances under both evidence conditions, median self-reported confidence remains 0.9 in both regimes despite a 3$\times$ reduction in visual information. We corroborate this finding with logprob-derived confidence ($p_{\max}$), obtained via a separate prompt interface on matched question instances; this signal exhibits the same failure mode. The model does not ``know when it does not know'' under shift. These results motivate warrant-based selective prediction, where confidence is explicitly bounded by what the available evidence can support.
\end{abstract}

\section{Introduction}

\subsection{Motivation}

Vision-language models (VLMs) are increasingly deployed for tasks requiring interpretation of visual information in context, from medical image analysis to autonomous vehicle decision-making. In such high-stakes applications, systems must have the ability to abstain when uncertain rather than risk costly errors. This capability, known as \emph{selective prediction} \citep{geifman2017selective,el2010foundations}, allows trading coverage (fraction of inputs answered) for reduced error rates among predictions that are made.

Modern VLMs typically provide confidence scores alongside predictions. A natural approach to selective prediction is \emph{confidence-based abstention}, where predictions are accepted only when confidence exceeds a threshold $\varepsilon$. The central question is whether model-reported confidence provides reliable epistemic information about prediction quality.

For confidence to support reliable selective prediction, two properties are necessary:
\begin{enumerate}
    \item \textbf{In-distribution control}: Sweeping $\varepsilon$ should yield smooth, monotone risk-coverage tradeoffs. Higher thresholds should reliably reduce error rates among accepted predictions.
    \item \textbf{Robustness to evidence shifts}: When input quality degrades (e.g., fewer frames in video, lower resolution), confidence should decrease accordingly. The model should ``know when it does not know'' due to insufficient evidence.
\end{enumerate}

The first property is often demonstrated in practice; the second remains poorly understood. If confidence is insensitive to evidence quality, confidence-based gates provide a false sense of security. The system appears to abstain appropriately in-distribution, but confidence contracts insufficiently under degraded evidence conditions. Selectivity increases only modestly despite substantial information loss.

\subsection{This Work}

We evaluate confidence-based abstention for video question answering (VideoQA). VideoQA is well-suited for this study because evidence quality can be precisely controlled through frame sampling. We use NExT-QA \citep{nextqa} with Gemini 2.0 Flash to test whether confidence behaves mechanistically (smoothly monotone with threshold changes) and whether it remains calibrated under controlled evidence degradation.

Our contributions are:
\begin{enumerate}
    \item Confidence-based abstention provides \textbf{mechanistic control in-distribution}. Sweeping threshold $\varepsilon$ produces smooth, monotone risk-coverage curves with a clear operating regime (63.7\% coverage, 9.4\% error vs. 98.7\% coverage, 23.6\% error).
    
    \item This control is \textbf{not epistemic}. Evaluating the same frozen question instances under both 18-frame and 6-frame evidence conditions, the model's confidence distribution contracts insufficiently. Median self-reported confidence remains 0.9 in both regimes despite a 3$\times$ reduction in visual information. We corroborate this finding with logprob-derived confidence ($p_{\max}$), obtained via a separate prompt interface on matched question instances; this signal exhibits the same failure mode. Confidence does not track information availability.
    
    \item The gap between mechanistic control and epistemic validity motivates \textbf{warrant-based selective prediction}, where confidence is bounded by what the available evidence can support. Our results validate the need for such mechanisms but do not yet demonstrate their implementation.
\end{enumerate}

A system deployed under varying evidence conditions (e.g., intermittent video feeds, lossy compression) cannot rely on confidence thresholds tuned in-distribution. The threshold that achieves 9\% error at 63.7\% coverage on full evidence yields 9\% error at only 53.7\% coverage when evidence is degraded.

\section{Related Work}

We situate our work within several related areas: selective prediction, calibration under shift, conformal risk control, epistemic uncertainty, selective QA, multimodal uncertainty, mechanism design, and video QA benchmarks.

\subsection{Selective Prediction and Abstention}

Selective prediction (classification with a reject option) allows models to abstain when uncertain, trading coverage for reduced error rates. The foundational work of \citet{chow1970optimum} established the optimal reject rule: a confidence threshold that minimizes misclassification risk given a cost for abstaining. \citet{el2010foundations} formalized the risk-coverage framework for selective classification, characterizing conditions under which softmax confidence yields near-optimal selective classifiers.

Modern deep learning has revived interest in selective prediction. \citet{geifman2017selective} demonstrated that softmax-based thresholding provides smooth risk-coverage tradeoffs for DNNs on image classification, achieving 2\% top-5 error on ImageNet at 60\% coverage. \citet{geifman2019selectivenet} proposed SelectiveNet, which jointly trains a classifier and rejection head to optimize coverage at a target error rate, rather than relying on pure confidence thresholding. \citet{franc2023optimal} derived optimal reject strategies in closed form and introduced the ``proper uncertainty score'' concept, proving these achieve the best possible error-coverage tradeoff for any given model.

Recent work addresses calibration requirements for selective prediction. \citet{fisch2022calibrated} showed that standard selective classifiers can be poorly calibrated on accepted predictions (``uncertain uncertainty''), and proposed methods ensuring that accepted predictions are well-calibrated. \citet{hendrycks2017baseline} established maximum softmax probability as the canonical baseline for detecting misclassified and out-of-distribution examples, directly relevant to confidence-based abstention.

\textbf{Where our work differs:} We are not proposing a new selector; we are diagnosing that even a clean selector knob can be non-epistemic under evidence loss.

\subsection{Selective Prediction Under Distribution Shift}

A critical question is whether selective classifiers remain reliable under distribution shift. \citet{liang2024selective} generalize selective classification to handle covariate and label shift, noting that traditional methods assumed i.i.d.\ data and introducing confidence scoring functions that improve reliability on shifted data. \citet{heng2025knowabstainoptimalselective} revisit selective prediction through the Neyman-Pearson lens, showing that the optimal acceptance rule is a likelihood-ratio test. They evaluate under covariate shift and propose new selection scores combining distance-from-training-data with model logits, finding improved robustness.

However, both works study \textbf{distributional perturbation} (domain adaptation, covariate corruption) rather than \textbf{information removal}. Our work introduces a distinct type of shift, evidence-completeness degradation via temporal subsampling.

\subsubsection{Evidence Truncation and Partial Observability}

Most selective prediction and calibration-under-shift work studies covariate shift, corruption shift, or domain adaptation, where inputs change but the underlying evidence may remain sufficient for the task. In contrast, our shift is an intervention on information availability. Temporal subsampling removes evidence about event order and persistence, which can reduce the Bayes-optimal predictability of many VideoQA questions. This places the setting closer to partial observability than to conventional corruption benchmarks. As a result, success requires not only monotone risk-coverage behavior under a fixed regime, but also that confidence contracts in response to reduced observability, a property not implied by standard selective classification or calibration results. Our experiments show that Gemini's confidence distribution shifts only modestly under a 3$\times$ reduction in frames, even when selective prediction remains monotone, motivating evidence-conditioned constraints rather than confidence-only gating.

\subsection{Calibration Under Distribution Shift}

Calibration requires that predicted probabilities reflect true outcome frequencies. \citet{guo2017calibration} showed that modern DNNs are often overconfident and that temperature scaling provides effective post-hoc calibration in-distribution. \citet{hendrycks2019benchmarking} introduced the ImageNet-C/P corruption benchmarks as a canonical setting for evaluating robustness under ``corruption shift,'' which we contrast against our ``evidence completeness shift.''

\citet{ovadia2019uncertainty} conducted a large-scale study of predictive uncertainty under dataset shift, finding that post-hoc calibration ``falls short'' while ensemble methods retain calibration across shifts. \citet{lakshminarayanan2017ensembles} showed that deep ensembles capture uncertainty effectively without Bayesian machinery, often outperforming single models under shift. \citet{gal2016dropout} introduced MC dropout as an approximate Bayesian uncertainty method. \citet{zou2023adaptive} proposed Adaptive Calibrator Ensemble (ACE), which trains calibrators on both in-distribution and challenging OOD data.

These studies focus on covariate shifts where the input distribution changes but the mapping from inputs to outputs remains valid. Our evidence degradation intervention is different. The input contains less information about the answer.

\subsection{Conformal Prediction and Risk Control}

Conformal prediction~\citep{vovk2022algorithmic,shafer2008tutorial} provides distribution-free coverage guarantees: prediction sets that contain the true label with probability at least $1-\alpha$, without distributional assumptions beyond exchangeability. \citet{angelopoulos2021gentle} provide an accessible tutorial showing conformal methods applied to modern deep learning.

The closest ``contract-like'' alternative to our warrant constraint is Conformal Risk Control. \citet{angelopoulos2024conformal} extend conformal prediction from coverage sets to controlling expected loss, with variants that address certain types of shift. \citet{xu2025selectiveconformalriskcontrol} combine selective classification with conformal prediction in Selective Conformal Risk Control (SCRC), a two-stage approach that first filters uncertain inputs, then constructs conformal sets for accepted inputs.

However, conformal methods control \textit{marginal} risk under exchangeability (or specified relaxations), not an explicit bound by information-theoretic observability. If we want assurance that \textit{this particular prediction} is reliable given \textit{this particular input's information content}, conformal methods do not directly provide it. Our warrant-based framing seeks per-instance guarantees where confidence is bounded by what the specific evidence can support.

\subsection{Epistemic Uncertainty and Information-Theoretic Bounds}

Epistemic uncertainty (model uncertainty due to limited knowledge) is distinct from aleatoric uncertainty (inherent randomness in data). Bayesian neural networks and Monte Carlo Dropout \citep{gal2016dropout} estimate epistemic uncertainty via posterior variance. \citet{lakshminarayanan2017ensembles} showed that deep ensembles capture uncertainty effectively without Bayesian machinery.

Information theory provides fundamental limits on predictability. Fano's inequality connects conditional entropy $H(Y|X)$ to minimum achievable error: if features contain limited information about labels, no classifier can be highly accurate \citep{fano1961transmission}. This implies an upper bound on justified confidence. \citet{sensoy2018evidential} operationalize this intuition through evidential deep learning, where models output Dirichlet distributions over classes. When evidence is weak, the Dirichlet is diffuse, yielding high-entropy predictions.

Our empirical evaluation tests whether VLM confidence actually respects such bounds in practice. When we reduce frame count, we reduce information about temporal events. A warrant-respecting model would lower confidence accordingly. We find that Gemini's confidence distribution contracts only modestly.

\subsection{Selective QA and Unanswerability}

The question of whether models ``know what they don't know'' has been studied extensively in question answering. \citet{rajpurkar2018squad2} introduced SQuAD 2.0, treating unanswerability as a first-class outcome rather than an error mode. \citet{kamath2020selective} studied selective QA under domain shift, showing that softmax-based abstention fails under OOD conditions and proposing a calibrator-based approach.

Most relevant to our work, \citet{whitehead2022reliable} introduced Reliable VQA, explicitly framing the problem as ``abstain rather than answer incorrectly.'' They use risk-coverage analysis in VQA and show that naive softmax thresholding can give extremely low coverage at low risk. Our work extends this evaluation style to video, with the novelty being the controlled evidence-completeness intervention.

\subsection{Uncertainty in Multimodal and VLM Systems}

Recent work has begun examining uncertainty and calibration specifically in vision-language models. \citet{oh2024calibrated} study calibrated robust fine-tuning of VLMs, showing that standard fine-tuning degrades OOD calibration and proposing methods to improve both OOD accuracy and calibration jointly. \citet{chen2024calibration} analyze calibration and uncertainty behavior in multimodal LLMs, introducing an ``I don't know'' evaluation dataset and finding that MLLMs tend to answer rather than admit uncertainty; they show that prompting strategies can improve self-assessment but do not eliminate miscalibration. \citet{wen2025abstention} provide a comprehensive survey of abstention in LLMs, offering a taxonomy that distinguishes behavioral refusal (alignment-driven) from epistemic abstention (uncertainty-driven), useful for positioning our ``warrant'' as an epistemic contract rather than a behavioral refusal.

These works establish that multimodal models often do not self-assess uncertainty well, motivating our empirical investigation of whether confidence tracks evidence quality in video understanding.

\subsection{Mechanism Design for Trustworthy Deployment}

Mechanism design offers an alternative perspective. Instead of training models to be calibrated, design incentive structures that make honest confidence reporting optimal. \citet{zhao2021mechanism} propose an insurance-based mechanism between forecasters and decision-makers, where the forecaster backs predictions with bets. Stakes are set so that truthful probability reporting is optimal, providing individual-level reliability guarantees.

Proper scoring rules \citep{gneiting2007scoring} formalize incentives for honest probability forecasts. A strictly proper scoring rule (like log-loss) ensures that reporting true beliefs maximizes expected score. This connects to our warrant concept. A warrant-respecting model effectively commits to a contract where confidence is bounded by evidence-supportable accuracy.

\subsection{Video Question Answering}

NExT-QA \citep{nextqa} provides a benchmark for temporal and causal reasoning in video, distinguishing descriptive questions (answerable from single frames) from temporal questions (requiring event ordering) and causal questions (requiring understanding of why events occur). This structure enables controlled evaluation of evidence requirements.

Modern VLMs including Gemini \citep{gemini2024} achieve strong performance on video understanding tasks, but systematic evaluation of their confidence behavior under evidence degradation is lacking. Prior VideoQA work focuses on accuracy metrics rather than confidence calibration or selective prediction. Our work fills this gap by treating frame sampling as a controlled intervention on evidence quality.

\subsection{Summary: Our Contribution}

Existing work on selective prediction and calibration studies how confidence thresholds trade coverage for accuracy under a fixed evidence regime, and conformal prediction provides marginal distribution-free guarantees under exchangeability. In contrast, our experiments isolate an intervention on observability that changes the evidence view itself. This motivates analyzing confidence as a function of the evidence available to support a claim, rather than as a purely model-internal score. We formalize this perspective through the warrant $\zeta(e)$, the Bayes-optimal predictability of a claim given an evidence view $e$, and study whether deployed confidence signals contract appropriately under evidence truncation.

To our knowledge, this is the first empirical study of whether VLM confidence tracks evidence completeness under controlled degradation. Prior selective prediction work assumes fixed distributions or tests distributional shifts (covariate corruption, domain shift) that do not isolate information content. Prior calibration work tests corruptions that degrade image quality but not necessarily information content. Conformal methods provide marginal guarantees under exchangeability, not per-instance bounds tied to observability. Our contribution is demonstrating that confidence-based abstention provides mechanical control in-distribution but fails to provide epistemic guarantees under evidence shifts, motivating warrant-based formulations.

\section{Experimental Setup}

\subsection{Dataset}

We use NExT-QA~\citep{nextqa}, a video question answering benchmark designed to evaluate temporal and causal reasoning. The dataset contains three question types. \textbf{Descriptive} questions ask about static attributes (what/who/where) that can be answered from individual frames. \textbf{Temporal} questions require understanding event ordering across time. \textbf{Causal} questions probe why or how events occur, requiring deeper reasoning about relationships between actions.

Each question has five multiple-choice options labeled A through E. We evaluate on 300 items from the validation split, stratified to include 100 questions of each type. This item list is frozen in \texttt{item\_ids.json} to ensure exact reproducibility across all experiments, including the Evidence Degradation conditions. For any cross-method comparison (self-reported vs.\ logprob-derived confidence), we match predictions by \texttt{question\_id} under the same evidence condition and report statistics on the intersection where both methods produce valid outputs ($n=295$ at 18 frames, $n=292$ at 6 frames).

\subsection{Evidence Packet Construction}

We extract a fixed set of frames from each video using a two-stage sampling strategy to ensure deterministic, reproducible inputs. First, we sample 12 frames uniformly across the video's full duration $T$, placing frame $i$ at timestamp $t_i = (i + 0.5) / 12 \cdot T$ for $i \in \{0, \ldots, 11\}$. This provides broad temporal coverage. Second, we extract 6 additional frames from the middle third of the video (between 33\% and 66\% of the duration) at timestamps $t_j = (0.33 + (j + 0.5) / 6 \cdot 0.33) \cdot T$ for $j \in \{0, \ldots, 5\}$. This ``zoom'' region focuses on the temporal center where key events often occur in short video clips.

\begin{figure}[t]
    \centering
    \includegraphics[width=\textwidth]{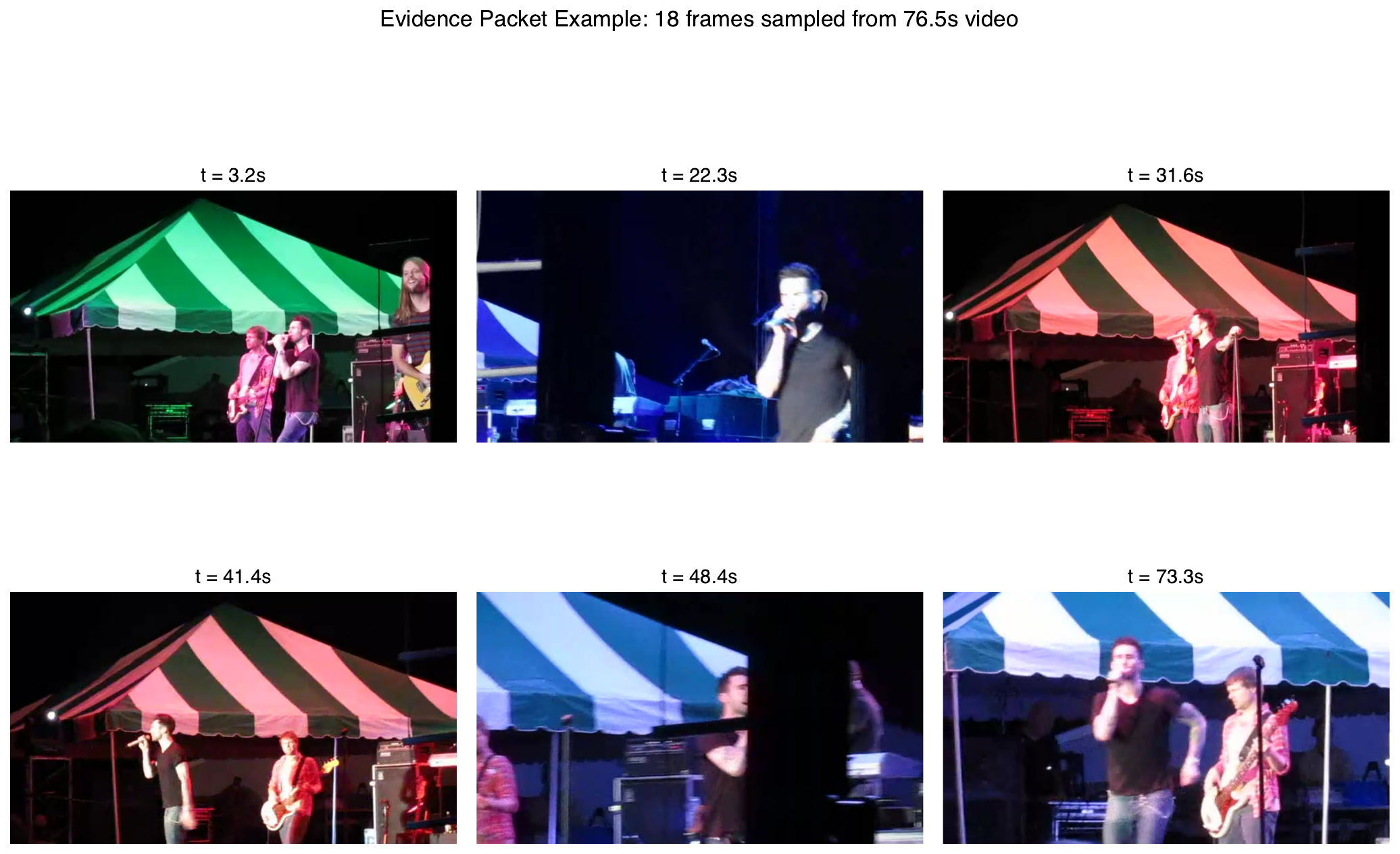}
    \caption{Example evidence packet showing 6 of 18 frames extracted from a 76.5-second video. Frames are sampled to provide both broad temporal coverage (uniform sampling) and focus on the middle third (zoom region). Timestamps are shown above each frame.}
    \label{fig:frame-examples}
\end{figure}

After merging both sets of timestamps, we sort and deduplicate any frames within 150ms of each other, typically yielding 15--18 frames per video. Each frame is extracted at its precise timestamp, resized to 512 pixels on the short side while preserving aspect ratio, and encoded as a JPEG with quality 85. The resulting frames are stored alongside a manifest file containing the extraction parameters, frame timestamps, and SHA256 hashes of each image. This cryptographic verification ensures that every experimental run uses identical visual inputs. Figure~\ref{fig:frame-examples} shows an example evidence packet with 6 representative frames from an 18-frame sequence.

\begin{table}[h]
\centering
\begin{tabular}{lcc}
\toprule
\textbf{Parameter} & \textbf{Baseline} & \textbf{Sparse (6f)} \\
\midrule
Uniform frames & 12 & 6 \\
Zoom frames & 6 & 0 \\
Typical total frames & 15--18 & 6 \\
JPEG quality & 85 & 85 \\
\bottomrule
\end{tabular}
\caption{Evidence packet parameters for baseline and shift conditions.}
\label{tab:evidence-params}
\end{table}

\subsection{Model and Prompting}

We use Gemini 2.0 Flash~\citep{gemini2024} as our vision-language model, configured with temperature 0 to ensure deterministic outputs and a maximum of 256 output tokens. For each question, the model receives the extracted frames in chronological order, the question text, and all five answer options labeled A through E. The prompt instructs the model to select one option or explicitly abstain if the visual evidence is insufficient.

The model is required to output a structured JSON object containing four fields. The \texttt{choice} field specifies the selected answer (A--E) or null if abstaining. The \texttt{confidence} field provides a numerical score in $[0, 1]$ representing the model's confidence in its answer. The \texttt{abstain} field is a boolean flag indicating whether the model chooses to abstain. Finally, the \texttt{evidence\_span} field identifies a contiguous range of frame indices that support the answer. The prompt forbids any free-form text or explanations, enforcing strict structured output that can be parsed reliably.

\subsection{System-Level Abstention}

We define abstention at the \emph{system level} based solely on a confidence threshold $\varepsilon$, rather than relying on the model's internal decision to abstain. This distinguishes our approach from prior work on model self-abstention. A prediction $p$ is considered abstaining if any of the following conditions hold: the JSON output failed to parse, the model returned a null choice, the confidence value is missing, or the confidence falls below threshold $\varepsilon$. Formally:
\begin{equation}
\text{abstain}_\text{sys}(p, \varepsilon) = \begin{cases}
\text{True} & \text{if parse failure or null prediction} \\
\text{True} & \text{if } p.\text{confidence} \text{ is missing} \\
\text{True} & \text{if } p.\text{confidence} < \varepsilon \\
\text{False} & \text{otherwise}
\end{cases}
\end{equation}
This formulation ensures that abstention behavior is fully controlled and auditable by the system designer. The model's self-reported \texttt{abstain} flag is logged for analysis but is not used for gating decisions. This design allows us to systematically sweep the threshold $\varepsilon$ and measure the resulting risk-coverage tradeoff without confounding effects from the model's alignment training.

\subsection{Evaluation Metrics}

We use standard selective prediction metrics~\citep{geifman2017selective}. Let $P$ denote the full set of predictions and $A_\varepsilon \subseteq P$ the subset accepted (not abstaining) at threshold $\varepsilon$. \textbf{Coverage} is the fraction of inputs answered, $\text{Coverage}(\varepsilon) = |A_\varepsilon| / |P|$. \textbf{Risk} is the error rate among accepted predictions, $\text{Risk}(\varepsilon) = \text{errors among } A_\varepsilon / |A_\varepsilon|$. These metrics characterize the fundamental tradeoff in selective prediction. Sweeping $\varepsilon$ from 0 to 1 traces a risk-coverage curve where lower coverage (more abstention) should yield lower risk (fewer errors among accepted predictions).

We also measure \textbf{calibration} using Expected Calibration Error (ECE)~\citep{guo2017calibration}, computed over the accepted predictions at each threshold:
\begin{equation}
\text{ECE} = \sum_{b=1}^{B} \frac{|B_b|}{n} \left| \text{acc}(B_b) - \text{conf}(B_b) \right|
\end{equation}
where predictions are partitioned into $B$ bins by confidence, $B_b$ is the set of predictions in bin $b$, $\text{acc}(B_b)$ is the fraction correct, and $\text{conf}(B_b)$ is the mean confidence. Well-calibrated predictions should have ECE near zero.

An important methodological note concerns statistical power. At extreme values of $\varepsilon$ where few predictions are accepted ($|A_\varepsilon| < 50$), risk estimates have high variance and should not be interpreted. We mark such points as NaN and omit them from our analysis and figures.

\subsection{Logprob-Derived Confidence}
\label{sec:logprob-method}

Self-reported confidence is a behavioral interface that may not reflect the model's token-level decision distribution. We obtain token log probabilities via the Vertex AI SDK, which exposes logprobs for Gemini 2.0 Flash (\texttt{gemini-2.0-flash-001}), to investigate whether the decoder's preference signal over answer options provides better signal.

For logprob extraction, we use a simplified prompt that requests only a single letter response (A--E) without JSON structure. The model is configured with \texttt{response\_logprobs=True} and \texttt{logprobs=20} to return the top-20 token candidates with their log probabilities. We extract the log probabilities for tokens corresponding to answer options A, B, C, D, E from the first generated token, matching exact single-character tokens. If an option does not appear in the top-20, we assign $\ell_i = -100$ (effectively $-\infty$); in practice, all five options consistently appear in the returned candidates.

From the raw log probabilities $\{\ell_A, \ell_B, \ell_C, \ell_D, \ell_E\}$, we compute normalized probabilities via softmax:
\begin{equation}
p_i = \frac{\exp(\ell_i)}{\sum_{j \in \{A,B,C,D,E\}} \exp(\ell_j)}
\end{equation}
This renormalization is necessary because the model's full vocabulary distribution includes tokens beyond A--E. By extracting only the A--E logprobs and renormalizing, we obtain a probability distribution over the answer space that represents the model's token-level preference among the five choices---the decoder's ``voting distribution'' over answers, independent of the self-reported confidence scalar.

We then derive three confidence metrics from this distribution:
\begin{enumerate}
    \item \textbf{Maximum probability} ($p_{\max}$): The probability of the most likely answer, $\max_i p_i$.
    \item \textbf{Margin}: The difference between the top two probabilities, $p_{\max} - p_{\text{second}}$.
    \item \textbf{Normalized entropy}: $H(p)/H_{\max}$, where $H(p) = -\sum_i p_i \log p_i$ and $H_{\max} = \log 5$.
\end{enumerate}

These logprob-derived metrics provide an alternative confidence signal that reflects the model's token-level decision distribution over answer options, rather than a self-reported scalar. Because the logprob experiment uses a different prompt interface (single-letter response, no JSON structure), absolute accuracy values should not be compared directly with the self-reported confidence experiments; the relevant comparison is \emph{differences across evidence conditions} within each method.

Importantly, self-reported confidence and logprob-derived confidence are obtained via \emph{separate model calls} with different output interfaces (JSON vs.\ single-letter response with logprobs enabled). This can change instruction-following behavior and therefore absolute accuracy, so we avoid comparing cross-interface accuracy levels. Our comparisons focus on (i)~how each confidence signal changes under evidence degradation (18 $\to$ 6 frames) \emph{within} the same interface, and (ii)~whether the qualitative failure mode---confidence not contracting under evidence loss---persists across interfaces. When we juxtapose the two confidence signals in a single figure or table, we compute statistics on matched question instances (same \texttt{question\_id}, same evidence condition), restricted to the intersection where both calls produce valid outputs.

\section{Experiments}

\subsection{Pipeline Validation}

Before conducting large-scale experiments, we validate the end-to-end pipeline on a small sample of 50 randomly selected items from the validation set. We allow one retry for JSON parse failures. The model successfully parses all 50 outputs (100\% success rate), achieves 64\% baseline accuracy, and averages 7.5 seconds per query. These results confirm that the frame extraction, API communication, JSON parsing, and evaluation components function correctly.

\subsection{Baseline Risk-Coverage}

This experiment tests whether confidence-based abstention provides mechanistic control over the risk-coverage tradeoff. We process all 300 items in our frozen validation set (100 causal, 100 temporal, 100 descriptive), querying the model once per item. Temperature is set to 0 for determinism. Parse failures are retried once; if the retry fails, the prediction is treated as an abstention.

We sweep the confidence threshold $\varepsilon$ over 25 evenly spaced values in $[0, 1]$. At each threshold, we compute coverage (fraction of items answered), risk (error rate among answered items), and Expected Calibration Error among accepted predictions. This produces risk-coverage curves that characterize the fundamental tradeoff in selective prediction. We also generate reliability diagrams at representative operating points to assess calibration.

A successful result exhibits four properties: (1) a smooth risk-coverage curve with a visible ``knee'' where modest coverage reductions yield substantial risk reductions; (2) monotonic variation in both risk and abstention rate with $\varepsilon$; (3) improved calibration (lower ECE) at higher thresholds.

\subsection{Evidence Degradation}

\begin{figure}[t]
    \centering
    \includegraphics[width=\textwidth]{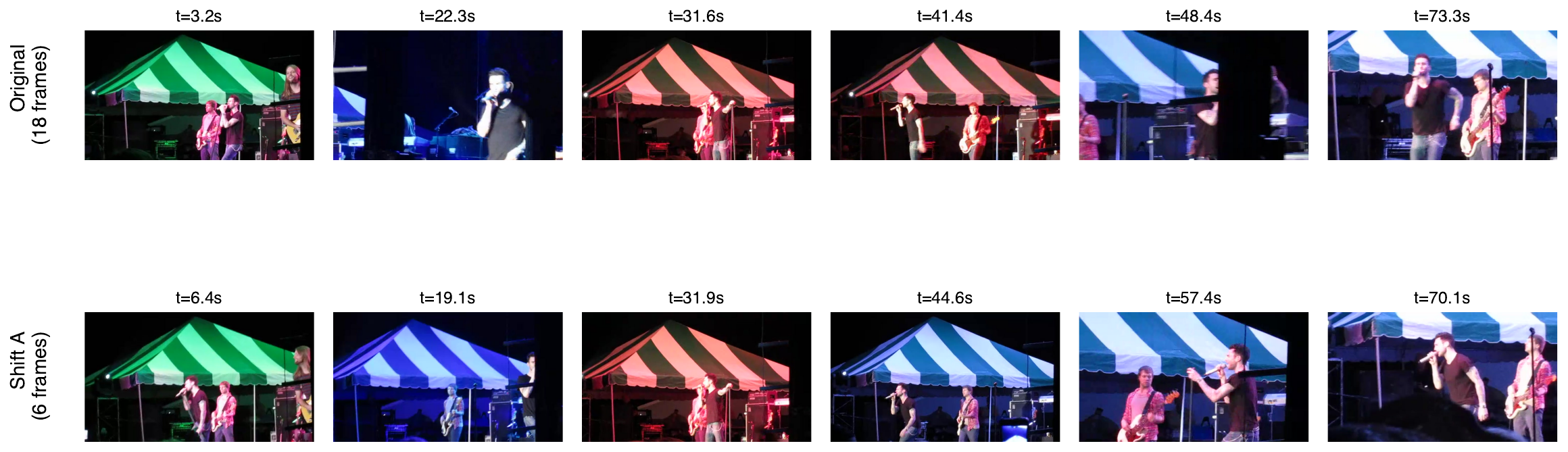}
    \caption{Visual comparison of evidence packets for the same video. Top row shows 6 frames sampled from the original 18-frame evidence packet. Bottom row shows all 6 frames from the degraded condition (Sparse). The degraded condition provides much sparser temporal coverage of the video's 76.5-second duration.}
    \label{fig:evidence-comparison}
\end{figure}

This experiment tests whether confidence remains calibrated under distribution shift. We evaluate the model on the same 300 questions but with degraded visual evidence: only 6 uniformly sampled frames and no zoom frames, reducing the typical frame count from 15--18 down to 6. This simulates compromised temporal resolution while spatial resolution remains unchanged. Figure~\ref{fig:evidence-comparison} illustrates the difference between original and degraded evidence for the same video.

We rerun the full selective prediction pipeline with the degraded evidence and sweep $\varepsilon$ over the same 25 values as in Baseline Risk-Coverage. Comparing the risk-coverage curves between original and degraded conditions reveals whether the model's confidence tracks evidence quality. A well-calibrated model should report lower confidence when given fewer frames, leading to decreased coverage at any fixed threshold. Additionally, the risk-coverage curve should remain smooth and monotone, though potentially shifted. The degradation should be most pronounced for temporal questions, which depend more heavily on observing events across time.

\subsection{Logprob Confidence}

Self-reported confidence may not reflect the model's token-level decision distribution. We conduct an additional experiment using logprob-derived confidence scores (Section~\ref{sec:logprob-method}) to test whether the decoder's preference signal over answer options provides better signal for selective prediction and, in particular, whether it is more sensitive to evidence degradation.

We query the same 300 items across four evidence conditions: original (18 frames), Sparse (6 frames), early-half (first 50\% of video), and late-half (second 50\% of video). For each query, we extract the token-level probability distribution over answer options and compute three confidence metrics: maximum probability ($p_{\max}$), margin, and normalized entropy.

We test whether logprob-derived confidence shows greater sensitivity to evidence degradation than self-reported confidence. If the model's token-level decision distribution tracks information availability, we expect: (1) lower $p_{\max}$ values when evidence is degraded, (2) smaller margin between top options under uncertainty, and (3) higher entropy when the model cannot discriminate between answers. If both confidence signals show similar insensitivity to evidence reduction, the overconfidence problem is fundamental to the model's representations rather than an artifact of the self-reporting interface.

\section{Results}

\subsection{Baseline Risk-Coverage}

Table~\ref{tab:baseline-results} summarizes results at five representative operating points. The model successfully parsed 297 of 300 queries (99\% success rate), with the 3 parse failures treated as abstentions. At the baseline threshold of $\varepsilon=0$ (accepting all valid predictions), the system achieves 98.7\% coverage with 23.6\% risk. As the threshold increases, coverage decreases while risk among accepted predictions drops substantially.

\begin{table}[h]
\centering
\begin{tabular}{lccccc}
\toprule
 & \textbf{$\varepsilon=0$} & \textbf{$\varepsilon=0.54$} & \textbf{$\varepsilon=0.71$} & \textbf{$\varepsilon=0.83$} & \textbf{$\varepsilon=0.92$} \\
\midrule
Coverage & 98.7\% & 97.3\% & 63.7\% & 63.0\% & 17.3\% \\
Risk & 23.6\% & 22.6\% & 9.4\% & 9.0\% & 1.9\% \\
ECE & 0.067 & 0.062 & 0.018 & 0.015 & 0.009 \\
Accepted (n) & 296 & 292 & 191 & 189 & 52 \\
\bottomrule
\end{tabular}
\caption{Baseline Risk-Coverage results at selected operating points. Parse success was 99\% (297/300).}
\label{tab:baseline-results}
\end{table}

\subsubsection{Risk-Coverage Tradeoff}

The risk-coverage curve in Figure~\ref{fig:risk-coverage} exhibits the desired mechanistic properties. The curve is smooth and monotone, with a clear ``knee'' around 60--70\% coverage. Tightening $\varepsilon$ from 0 to 0.71 reduces coverage from 98.7\% to 63.7\% (a 35 percentage point drop) while reducing risk from 23.6\% to 9.4\% (a 60\% relative reduction in error rate). Modest sacrifices in coverage yield substantial gains in reliability. The curve continues to improve at higher thresholds, though with diminishing returns and reduced sample sizes.

The monotonicity of this tradeoff is critical. Risk decreases smoothly as $\varepsilon$ increases, with no reversals or significant irregularities. The confidence signal correlates with correctness, so the abstention threshold provides predictable control over system behavior.

\subsubsection{Calibration Analysis}

\begin{figure}[t]
    \centering
    \begin{subfigure}[b]{0.32\textwidth}
        \includegraphics[width=\textwidth]{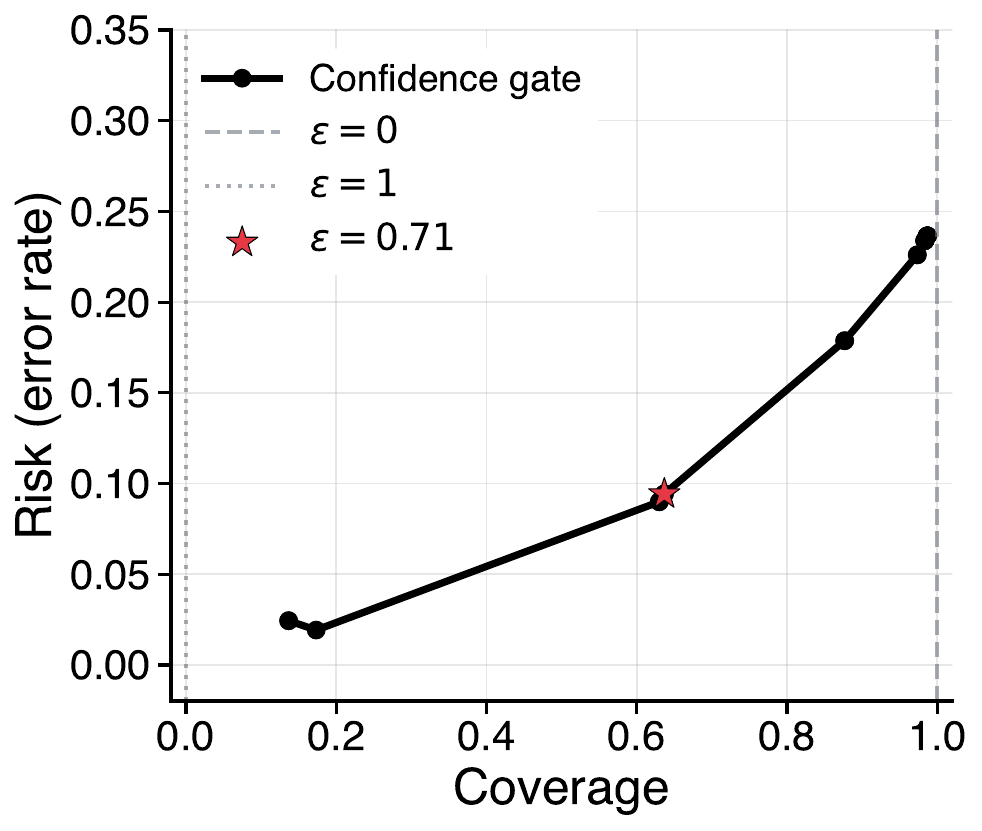}
        \caption{Risk-Coverage curve}
        \label{fig:risk-coverage}
    \end{subfigure}
    \hfill
    \begin{subfigure}[b]{0.32\textwidth}
        \includegraphics[width=\textwidth]{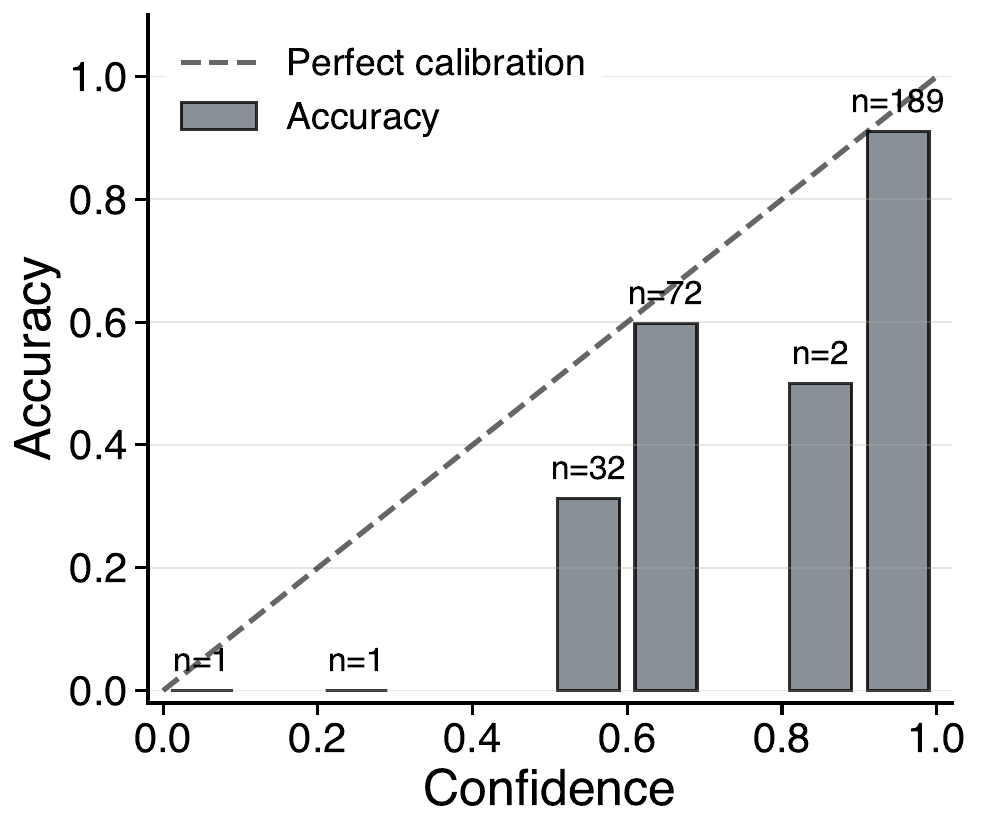}
        \caption{Reliability diagram}
        \label{fig:calibration}
    \end{subfigure}
    \hfill
    \begin{subfigure}[b]{0.32\textwidth}
        \includegraphics[width=\textwidth]{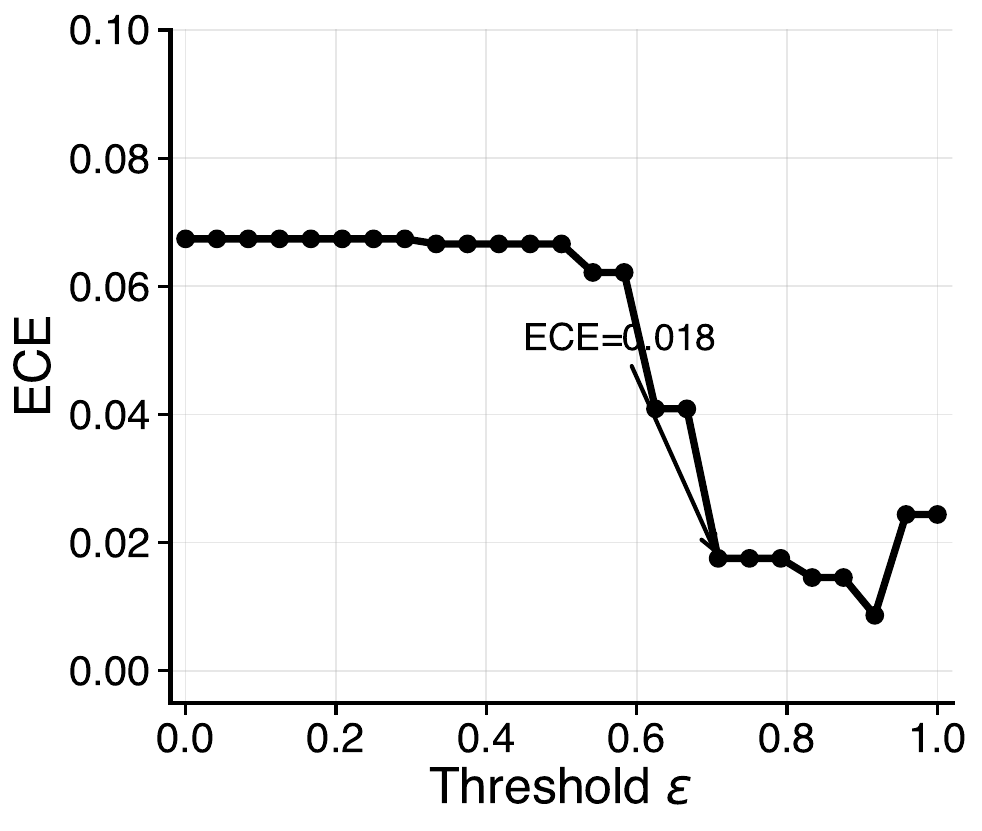}
        \caption{ECE vs threshold}
        \label{fig:ece}
    \end{subfigure}
    \caption{\textbf{Takeaway: In-distribution, confidence-based abstention works.} Baseline performance on 18 frames ($n=300$). (a)~\textbf{Risk-Coverage}: sweeping $\varepsilon$ produces a smooth, monotone curve---higher thresholds reliably reduce error. Starred point: $\varepsilon=0.71$ achieves 9.4\% risk at 63.7\% coverage. (b)~\textbf{Reliability diagram} (all predictions): bars show accuracy per confidence bin. The model is well-calibrated---predicted confidence matches actual accuracy. Most predictions cluster at high confidence (0.9--1.0 bin has $n=189$). (c)~\textbf{Expected Calibration Error (ECE) vs.\ threshold}: calibration error drops from 0.067 to 0.018 as we become more selective.}
    \label{fig:baseline-results}
\end{figure}

Calibration improves markedly as the confidence threshold tightens. Expected Calibration Error drops from 0.067 at $\varepsilon=0$ to 0.018 at $\varepsilon=0.71$ (Figure~\ref{fig:ece}). The reliability diagram at this operating point (Figure~\ref{fig:calibration}) shows strong calibration among high-confidence predictions. The 0.9--1.0 confidence bin contains 189 predictions with 91\% actual accuracy, demonstrating near-perfect alignment between reported confidence and empirical performance. The confidence gate successfully filters out poorly-calibrated low-confidence predictions, leaving well-calibrated high-confidence answers.

\subsection{Evidence Degradation}

Table~\ref{tab:degradation-results} compares performance between the original evidence (15--18 frames) and the degraded condition (6 frames). At the baseline threshold $\varepsilon=0$, risk increases from 23.6\% to 27.4\% under degraded evidence, a modest but noticeable degradation. At the fixed threshold $\varepsilon=0.71$, coverage decreases from 63.7\% to 53.7\%, and conditional risk remains similar (9.4\% vs 9.3\%). This pattern holds at all operating points: at the same epsilon, 6-frame predictions have lower coverage but similar conditional risk. The model does become more selective under degradation, but not selectively enough. At $\varepsilon=0.625$, where 18-frame achieves 87.7\% coverage with 17.9\% risk, 6-frame achieves 78.7\% coverage with 18.2\% risk. Despite having only one-third the visual information, the model's confidence distribution shifts only modestly.

\begin{figure}[t]
    \centering
    \includegraphics[width=\textwidth]{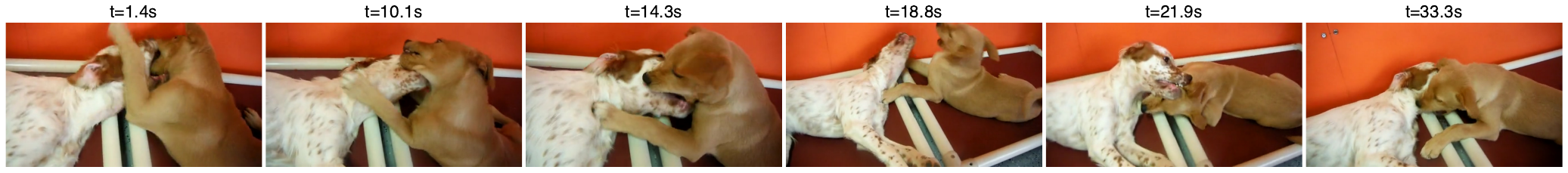}
    \caption{Overconfidence under evidence degradation. \textbf{Question:} ``How does the brown dog keep the white dog down?'' \textbf{Options:} A) change hand, B) hold the dog with its paws, C) walks around, D) using leash, E) lie down on chair. \textbf{Answer: B.} With 18 frames, the model answers correctly (B) with confidence 1.00. With only 6 frames (Sparse condition), the model answers incorrectly (A) yet still reports confidence 0.70. The question requires observing sustained behavior across time; sparse sampling misses critical moments, but the model fails to recognize that its evidence is insufficient.}
    \label{fig:question-example}
\end{figure}

Figure~\ref{fig:question-example} illustrates a concrete instance of this failure. The question asks ``how does the brown dog keep the white dog down,'' requiring observation of sustained interaction across time. With 18 frames spanning the full video, the model correctly identifies that the brown dog uses its paws (answer B) with confidence 1.00. With only 6 frames, the model misses critical moments and incorrectly answers A, yet reports confidence 0.70. The model does not recognize that the sparse sampling provides insufficient evidence.

\begin{table}[h]
\centering
\begin{tabular}{lcccc}
\toprule
\textbf{Condition} & \textbf{Risk @ $\varepsilon=0$} & \textbf{Risk @ $\varepsilon=0.71$} & \textbf{Coverage @ $\varepsilon=0.71$} & \textbf{$n_{\text{acc}}$} \\
\midrule
Original (18 frames) & 23.6\% & 9.4\% & 63.7\% & 191 \\
Sparse (6 frames) & 27.4\% & 9.3\% & 53.7\% & 161 \\
\bottomrule
\end{tabular}
\caption{\small Evidence Degradation results comparing original vs shifted evidence. At fixed $\varepsilon=0.71$, both conditions achieve similar conditional risk ($\sim$9\%), but 6-frame predictions have lower coverage (53.7\% vs 63.7\%), indicating the model is more selective but not by enough to compensate for the 3$\times$ evidence reduction.}
\label{tab:degradation-results}
\end{table}

\subsubsection{The Overconfidence Problem}

Overconfidence here refers to confidence persistence under reduced observability, not increased error at fixed confidence---Table~\ref{tab:degradation-results} shows conditional risk stays similar at fixed $\varepsilon$.

\begin{figure}[t]
    \centering
    \includegraphics[width=\textwidth]{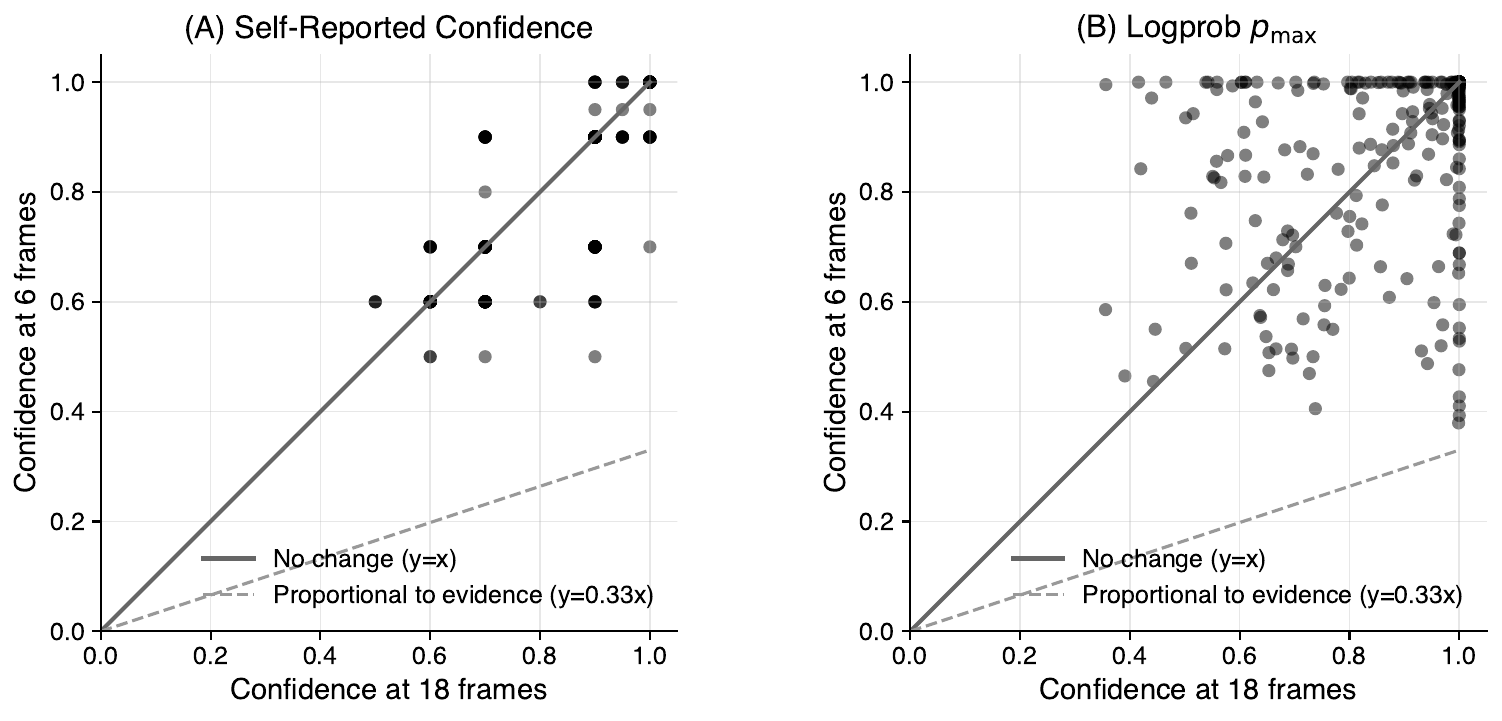}
    \caption{Per-item confidence comparison under evidence degradation (18 $\to$ 6 frames) on matched question instances. Each point is one question; both panels show the same $n=292$ items. Panel (A) appears sparser because self-reported confidence is discretized (0.6, 0.7, 0.9, 1.0, etc.), causing points to stack---darker regions indicate more overlapping points. If confidence tracked evidence, points would fall toward the dashed line (proportional 67\% reduction). Instead, points cluster near the diagonal, showing confidence is insensitive to evidence quality.}
    \label{fig:delta-summary}
\end{figure}

Figure~\ref{fig:delta-summary} captures the core finding directly: a 67\% reduction in evidence produces less than 6\% change in confidence. Self-reported confidence drops only 3.3\%, and logprob-derived $p_{\max}$ actually \emph{increases} by 0.6\%. The gap between evidence reduction and confidence reduction is the problem. A well-calibrated epistemic system would show commensurate contraction; instead, confidence remains high despite reduced observability.

\begin{figure}[t]
    \centering
    \includegraphics[width=\textwidth]{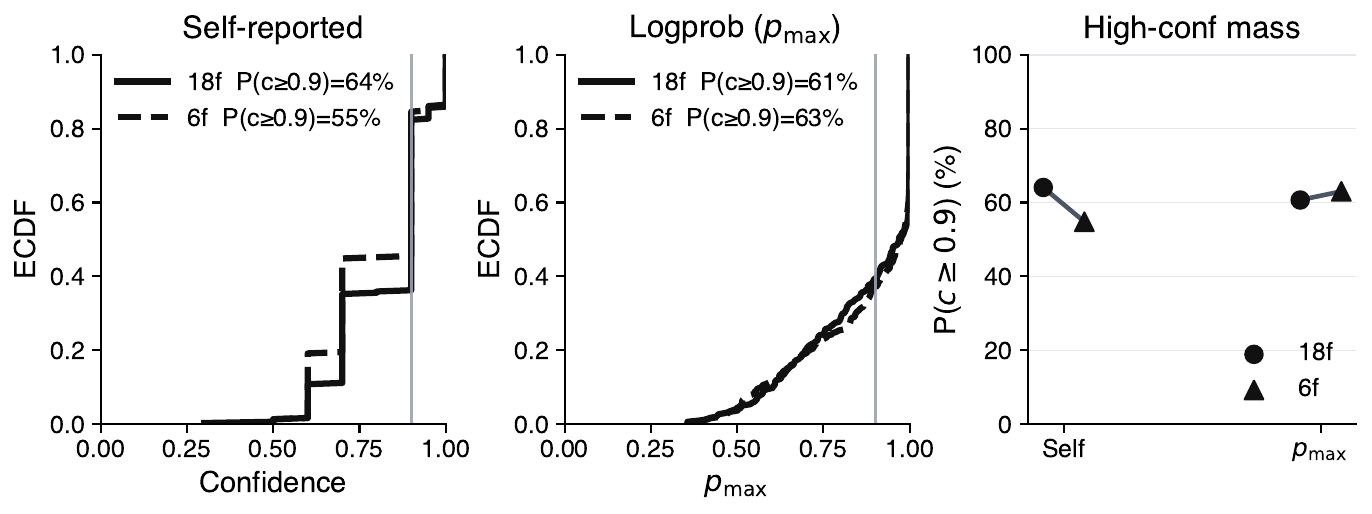}
    \caption{\textbf{Takeaway:} Even with 67\% less evidence, high-confidence mass barely decreases---and $p_{\max}$ actually increases---on the same questions. Left/middle: CDFs with a fixed high-confidence threshold ($c \ge 0.9$). Right: paired-dots summary of high-confidence mass. Self-reported and logprob confidence are computed on matched question instances (same \texttt{question\_id}, same evidence condition) via separate model calls; statistics use the matched intersection where both outputs are valid ($n=295$ at 18 frames, $n=292$ at 6 frames).}
    \label{fig:confidence-distribution}
\end{figure}

Figure~\ref{fig:confidence-distribution} shows this directly in the confidence signals on matched question instances. Despite a 3$\times$ reduction in visual information, the confidence CDFs shift only modestly. At a fixed high-confidence threshold ($c \ge 0.9$), self-reported confidence drops from 64\% to 55\%, while logprob-derived $p_{\max}$ increases from 61\% to 63\%. The model does not ``know when it does not know.''

\begin{figure}[h]
    \centering
    \includegraphics[width=0.7\textwidth]{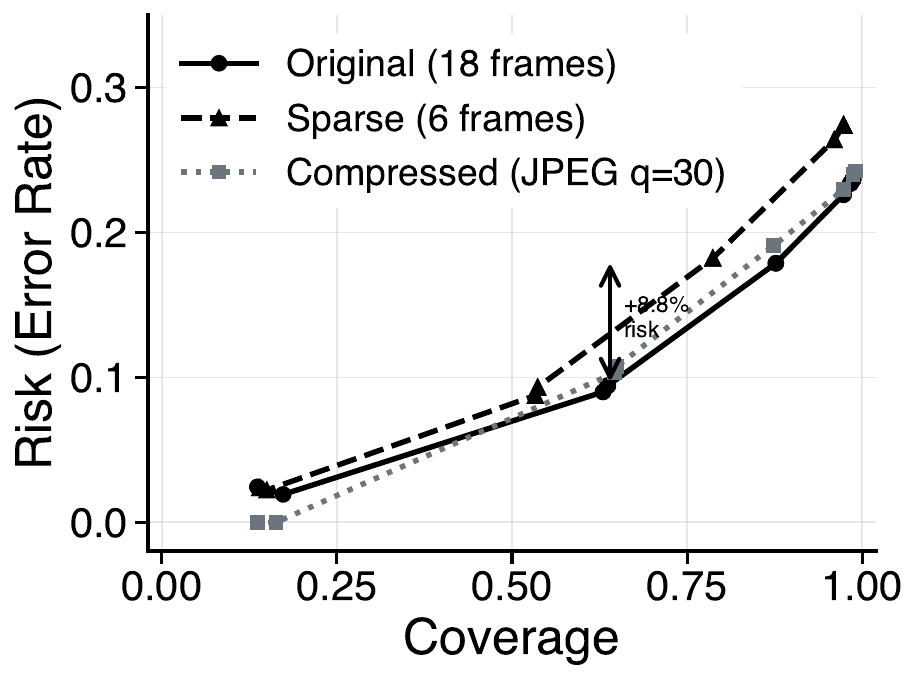}
    \caption{\small Risk-Coverage comparison under evidence degradation. Coverage is the fraction of predictions accepted (not abstaining); Risk is the error rate among accepted predictions. The Sparse curve (6 frames, dashed) shifts leftward and upward from the original (18 frames, solid). At any fixed $\varepsilon$, the 6-frame regime has lower coverage but similar conditional risk. Compressed (dotted) has minimal impact.}
    \label{fig:comparison}
\end{figure}

Figure~\ref{fig:comparison} shows the risk-coverage curves. At any fixed epsilon, the 6-frame regime has lower coverage but similar conditional risk. The model does become more selective, but not proportionally to the evidence reduction. At $\varepsilon=0.625$, the 18-frame regime achieves 87.7\% coverage with 17.9\% risk, while the 6-frame regime achieves 78.7\% coverage with 18.2\% risk. The coverage drops by only 9 percentage points despite a 3$\times$ reduction in frames.

The abstention mechanism still exhibits monotone behavior under shift. Tightening $\varepsilon$ continues to reduce risk, and the curve remains smooth without catastrophic failures. However, achieving the same risk level as the original condition requires much more aggressive thresholding, dramatically reducing coverage. Model confidence does not track evidence quality as an epistemic quantity. It appears to reflect task difficulty or other factors insensitive to visual input completeness.

\subsubsection{Diagnosing the Failure Mode}

The key diagnostic is illustrated in Figure~\ref{fig:comparison}. At any fixed epsilon, the 6-frame regime achieves lower coverage with similar conditional risk. At $\varepsilon=0.71$, coverage drops from 63.7\% to 53.7\% while risk remains at $\sim$9\%. At $\varepsilon=0.625$, coverage drops from 87.7\% to 78.7\% while risk stays at $\sim$18\%. The model is more selective under degradation, but the coverage reduction (10--15\%) is far smaller than the evidence reduction (67\%). The confidence distribution contracts, but not proportionally to the information loss.

As a control, we tested image compression by re-encoding all frames at JPEG quality 30 (versus baseline quality 85). This degradation had minimal impact. Risk at $\varepsilon=0$ increased by only 0.6 percentage points (24.2\% vs 23.6\%), and at 70\% coverage by 1.4 percentage points (10.8\% vs 9.4\%). The model is robust to compression artifacts but highly sensitive to temporal resolution. Frame count matters far more than image quality for video question answering.

\subsection{Logprob Confidence}

We compare risk-coverage curves using three logprob-derived confidence metrics: $p_{\max}$ (maximum softmax probability over answer options), margin (difference between top two probabilities), and normalized entropy. This tests whether the model's token-level decision distribution provides better signal than self-reported confidence.

Table~\ref{tab:logprob-results} summarizes logprob-derived confidence metrics across all four evidence conditions. The results reveal a striking pattern: logprob-derived confidence shows \emph{even less} sensitivity to evidence degradation than self-reported confidence.

\begin{table}[h]
\centering
\small
\begin{tabular}{lccccc}
\toprule
\textbf{Condition} & \makecell{\textbf{Acc ($\varepsilon=0$)} \\ \textit{\scriptsize no gating; logprob prompt}} & \textbf{Mean $p_{\max}$} & \textbf{Median $p_{\max}$} & \textbf{Mean Margin} & \textbf{Mean Entropy} \\
\midrule
Original (18 frames) & 82.4\% & 0.871 & 0.970 & 0.763 & 0.298 \\
Sparse (6 frames) & 81.3\% & 0.876 & 0.974 & 0.771 & 0.282 \\
Early-half (6 frames) & 77.3\% & 0.861 & 0.961 & 0.744 & 0.319 \\
Late-half (6 frames) & 78.7\% & 0.872 & 0.979 & 0.772 & 0.302 \\
\bottomrule
\end{tabular}
\caption{Logprob-derived confidence metrics across evidence conditions. Accuracy is unconditional (no gating, $\varepsilon=0$). Despite accuracy dropping from 82.4\% to 77--81\% under degradation, $p_{\max}$ remains remarkably stable (0.86--0.88) and median $p_{\max}$ stays above 0.96 in all conditions. Entropy remains in the narrow range 0.28--0.32 across all conditions despite 67\% evidence reduction.}
\label{tab:logprob-results}
\end{table}

\subsubsection{Comparison with Self-Reported Confidence}

\begin{figure}[h]
    \centering
    \includegraphics[width=\textwidth]{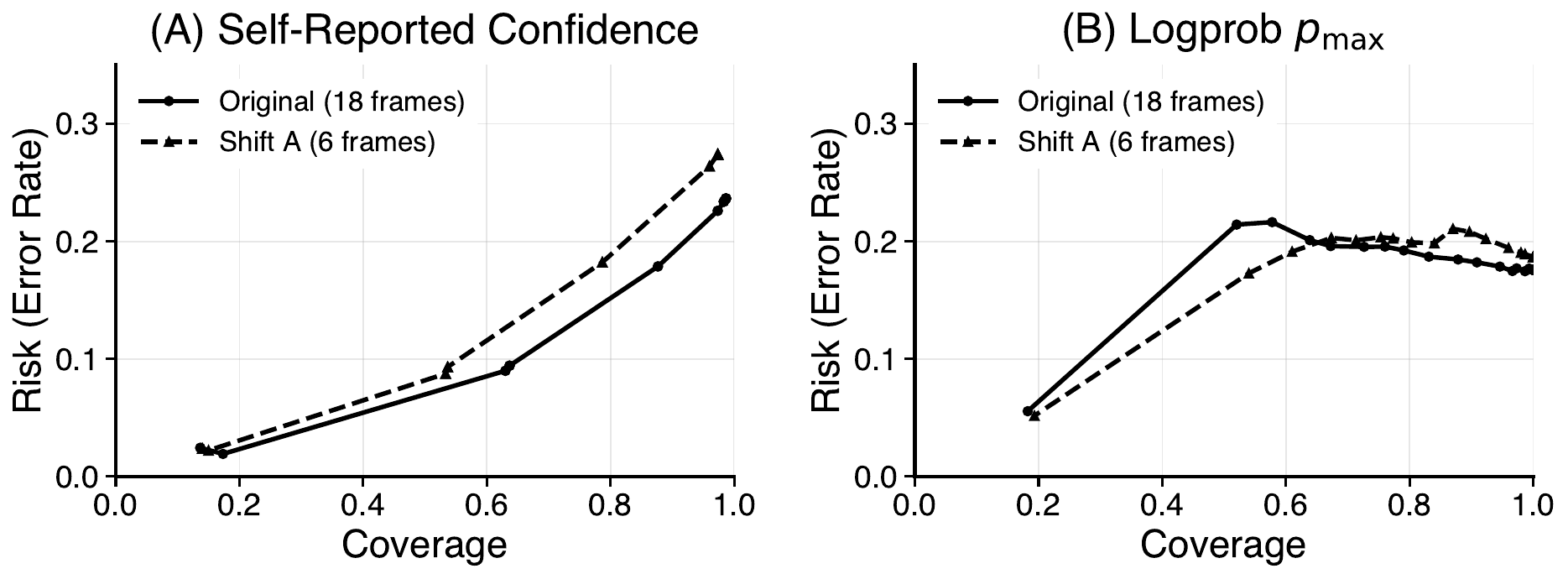}
    \caption{Risk-coverage curves comparing self-reported confidence (left) and logprob-derived $p_{\max}$ (right) under evidence degradation. Computed on matched question instances via separate model calls. Both methods exhibit leftward shift when frames are reduced from 18 to 6. However, the gap between conditions is \emph{smaller} for logprob confidence, indicating that the model's token-level decision distribution is even less sensitive to evidence quality than self-reported confidence.}
    \label{fig:confidence-comparison}
\end{figure}

Table~\ref{tab:confidence-comparison} directly compares self-reported confidence (from JSON output) with logprob-derived $p_{\max}$ on matched question instances (same \texttt{question\_id}, same evidence condition) via separate model calls. The comparison reveals that logprob confidence is systematically higher and less responsive to evidence degradation. Figure~\ref{fig:confidence-comparison} visualizes the risk-coverage curves for both confidence methods under evidence degradation.

\begin{table}[h]
\centering
\small
\begin{tabular}{lcccc}
\toprule
\textbf{Metric} & \multicolumn{2}{c}{\textbf{Self-Reported}} & \multicolumn{2}{c}{\textbf{Logprob $p_{\max}$}} \\
\cmidrule(lr){2-3} \cmidrule(lr){4-5}
 & 18 frames & 6 frames & 18 frames & 6 frames \\
\midrule
Mean & 0.832 & 0.804 & 0.870 & 0.876 \\
Median & 0.900 & 0.900 & 0.969 & 0.975 \\
$\Delta$ (degradation) & \multicolumn{2}{c}{$-$3.3\%} & \multicolumn{2}{c}{$+$0.6\%} \\
\bottomrule
\end{tabular}
\caption{Self-reported vs.\ logprob-derived confidence under evidence degradation. Computed on matched question instances (same \texttt{question\_id}, same evidence condition) via separate model calls; statistics use the matched intersection where both outputs are valid ($n=295$ at 18 frames, $n=292$ at 6 frames). Self-reported confidence drops 3.3\% (0.832 $\to$ 0.804) when frames are reduced. Logprob $p_{\max}$ actually \emph{increases} slightly (0.870 $\to$ 0.876). The model's token-level decision distribution is even less sensitive to evidence quality than its self-reported confidence.}
\label{tab:confidence-comparison}
\end{table}

\subsubsection{Interpretation}

The logprob results establish that overconfidence under evidence degradation is \textbf{not an artifact of the self-reporting interface}. If the problem were merely that the model's self-reported confidence diverged from its internal uncertainty, we would expect logprob-derived metrics to show greater sensitivity to evidence reduction. Instead, we observe the opposite:

\begin{itemize}
    \item Self-reported confidence drops 3.3\% under degradation (0.832 $\to$ 0.804)
    \item Logprob $p_{\max}$ \emph{increases} 0.6\% (0.870 $\to$ 0.876)
    \item Margin remains stable at 0.74--0.77 across all conditions
    \item Entropy remains in the narrow range 0.28--0.32 despite 67\% evidence reduction
\end{itemize}

This finding has important implications. The model's \emph{token-level decision distribution} over A--E does not become more diffuse when evidence is truncated. Both self-reported confidence and logprob-derived scores fail to track observability. The overconfidence problem is fundamental to the model's representations, not a behavioral artifact of the self-reporting interface.

The temporal ablation conditions (early-half and late-half) reinforce this conclusion. For questions whose answers depend on events in the missing temporal segment, these conditions provide semantically insufficient evidence by construction. Yet $p_{\max}$ remains above 0.86 and median $p_{\max}$ exceeds 0.96 in both cases. Because temporal ablation guarantees removal of one half of the clip, stability of $p_{\max}$ under early-half and late-half conditions suggests insensitivity is not explained solely by redundant sampling---the model genuinely does not recognize when critical temporal context is absent.

\section{Discussion}

\subsection{Summary of Findings}

Two main results emerge. First, confidence-based abstention provides mechanistic control over the risk-coverage tradeoff in-distribution. Sweeping threshold $\varepsilon$ from 0 to 0.71 reduces risk from 23.6\% to 9.4\% while maintaining 63.7\% coverage. The risk-coverage curve is smooth and monotone with a visible knee; calibration among accepted predictions is strong (ECE = 0.018). These are not artifacts of alignment training or stochastic variation. The abstention mechanism provides real, predictable control.

Second, this control does not transfer across distribution shift. When the number of frames is reduced from 18 to 6, the model's confidence distribution contracts only modestly (median confidence remains 0.9 in both regimes) despite a 3$\times$ reduction in visual information. At any fixed threshold, coverage decreases and conditional risk stays similar, but the coverage reduction (10--15\%) is far smaller than the evidence reduction (67\%). The model does not recognize when it has insufficient evidence to answer reliably.

Third, the logprob analysis confirms that overconfidence is fundamental to the model's representations, not an artifact of the self-reporting interface. Logprob-derived $p_{\max}$ actually \emph{increases} slightly under evidence degradation (0.870 $\to$ 0.876), while self-reported confidence at least decreases modestly (0.832 $\to$ 0.804). The model's token-level decision distribution over answer options does not become more diffuse when evidence quality degrades.

\subsection{Implications for Deployment}

A confidence threshold tuned on 15--18 frames does not preserve coverage when frame count drops. At $\varepsilon=0.71$, coverage falls from 63.7\% to 53.7\% while conditional risk stays near 9\%. Confidence gating remains valid for trading coverage against accuracy within a fixed evidence regime, but when input characteristics change (fewer frames, different sampling rates), the calibration no longer holds.

The shift is hard to detect from abstention rates alone because selectivity increases only modestly relative to the evidence reduction. Deployment monitoring cannot rely on coverage as a proxy for distribution shift. Instead, systems should track input characteristics directly: frame count, temporal coverage, motion density. These observability signals indicate when the model is operating outside its reliable regime.

The contrast between frame count and compression quality is also instructive. Reducing JPEG quality from 85 to 30 has minimal impact on accuracy (risk increases by only 1.4 percentage points), while reducing frame count from 18 to 6 increases risk from 23.6\% to 27.4\% at $\varepsilon=0$ and degrades the entire risk-coverage curve. This suggests that for video understanding tasks, maintaining temporal resolution is far more important than maintaining spatial resolution or image fidelity. Deployment systems should prioritize frame rate over bitrate.

\subsection{Implications for Warrant-Based Guarantees}

The results support two claims strongly but require careful interpretation regarding what they do and do not establish.

\subsubsection{What These Experiments Support}

First, \textbf{a control knob exists in-distribution}. Sweeping $\varepsilon$ yields monotone risk-coverage tradeoffs and improved calibration among accepted answers. The abstention mechanism provides predictable control for trading coverage against accuracy.

Second, \textbf{the control knob is not epistemic}. Under evidence degradation, confidence contracts insufficiently. Coverage drops modestly at fixed $\varepsilon$ (63.7\% to 53.7\%), but this 16\% reduction is far smaller than the 67\% evidence reduction. Median confidence remains 0.9 in both regimes. Confidence is not calibrated to information availability; it reflects correlates of task difficulty rather than evidential support.

Third, \textbf{the problem is representational, not behavioral}. Logprob-derived confidence ($p_{\max}$) shows even less sensitivity to evidence degradation than self-reported confidence, actually increasing slightly from 0.871 to 0.876 under frame reduction. This is inconsistent with the hypothesis that self-reported confidence merely diverges from internal uncertainty. The model's token-level probability distribution does not become more diffuse when evidence quality degrades.

These three facts imply a \emph{warrant-based} formulation in which reported confidence $p$ should satisfy $p \le \zeta(e) + \epsilon$ for some evidence-derived bound $\zeta(e)$. The Sparse condition failure is precisely the violation this contract would rule out. Confidence remains high when the evidence channel is weaker.

\subsubsection{What These Experiments Do Not Support}

These experiments do not yet validate that any proposed mechanism \emph{achieves} a warrant guarantee. We do not estimate a warrant quantity $\zeta$ (a measure of what the evidence supports), we do not produce a lower bound $\text{LB}(\zeta)$, and we do not enforce or audit the inequality $p \leq \text{LB}(\zeta)$.

What we have validated is the \emph{need} for such a guarantee and the \emph{inadequacy} of confidence-only gating. The experiments establish the problem statement, not the solution.

\subsubsection{Separating Selective Prediction from Warrant Guarantees}

These results separate ``selective prediction'' from the warrant guarantee. On the original evidence view (18 frames), a confidence threshold induces a clean risk-coverage tradeoff and improves calibration among accepted answers. Tightening $\varepsilon$ to 0.71 drops risk from 23.6\% to 9.4\% at 63.7\% coverage. Under evidence degradation (6 frames), the same threshold produces lower coverage (53.7\%) with similar conditional risk (9.3\%). The model is more selective, but the coverage reduction (16\%) is disproportionately small relative to the evidence reduction (67\%). The median confidence remains 0.9 in both regimes (Table~\ref{tab:warrant}), demonstrating that the model's subjective confidence does not contract with weaker evidence.

This is exactly the failure mode the warrant contract is meant to rule out. The contract constrains confidence relative to an evidence-conditioned warrant $\zeta(e)$, not relative to a distribution-specific calibration curve. The problem is that $p(S)$ behaves as if it were calibrated to correctness on one regime, but it violates the intended dominance condition $p(S) \leq \zeta(e) + \epsilon$ when $e$ changes, because confidence does not contract when the evidence view weakens.

Starting from the 6-frame regime and tuning $\varepsilon$ there looks ``safe'' when moving to 18 frames, but that is only a conservative policy selection. It does not constitute a guarantee, since the guarantee requires an explicit estimate or lower bound on $\zeta(e)$ and enforcement against that bound, not a threshold learned on one operating regime.

\subsubsection{Threshold Transfer Across Regimes}

We test the ``what if you started from shift?'' objection by computing threshold transfer in both directions. For each criterion (fixed risk or fixed coverage), we solve for $\varepsilon^*$ on the source regime via interpolation, then evaluate the same $\varepsilon^*$ on the target regime.

\paragraph{Fixed-Risk Transfer (Target: 10\% Risk)}

\begin{table}[h]
\centering
\small
\begin{tabular}{lcccccc}
\toprule
Direction & $\varepsilon^*$ & Source Risk & Source Cov & Target Risk & Target Cov & $n_{\text{acc}}$ \\
\midrule
18 $\to$ 6 & 0.706 & 9.4\% & 63.7\% (191) & 9.3\% & 53.7\% (161) & 191/161 \\
6 $\to$ 18 & 0.705 & 9.3\% & 53.7\% (161) & 9.4\% & 63.7\% (191) & 161/191 \\
\bottomrule
\end{tabular}
\caption{Fixed-risk transfer. The interpolated threshold achieves similar risk in both directions, but coverage differs substantially (63.7\% vs 53.7\%).}
\end{table}

\paragraph{Coverage Comparison at $\varepsilon=0.625$}

At $\varepsilon=0.625$ (the highest coverage operating point before the confidence threshold takes effect), we observe:
\begin{itemize}
    \item 18-frame: 87.7\% coverage, 17.9\% risk (263 accepted)
    \item 6-frame: 78.7\% coverage, 18.2\% risk (236 accepted)
\end{itemize}
Risk is nearly identical, but coverage differs by 9 percentage points. This means the 6-frame model is slightly more selective at any given epsilon, but the coverage reduction (9\%) is small relative to the evidence reduction (67\%). At matched epsilon, the model maintains similar calibration but admits fewer predictions under degradation. The problem is not that calibration breaks at fixed epsilon, but that the \emph{degree of selectivity increase} is insufficient for the \emph{degree of evidence loss}.

\subsubsection{Implications for Fine-Tuning}

The shift results reveal a missing capability: \textbf{confidence must become sensitive to evidence completeness}. Fine-tuning can learn this, but only with the right supervision signal, one tied to evidence quality rather than answer correctness alone.

A fine-tuning approach consistent with warrant-based guarantees would:
\begin{enumerate}
    \item Keep the same claim object (the multiple-choice answer)
    \item Add an auxiliary target derived from evidence availability, an \emph{observability proxy} such as frame count, temporal coverage, or motion magnitude
    \item Train so that reported confidence is \emph{monotone in evidence quality} and does not remain high when evidence is degraded
\end{enumerate}

However, fine-tuning alone does not create a guarantee. Fine-tuning improves the \emph{predictor} that feeds the contract; the guarantee comes from contract enforcement that gates predictions against a warrant-derived bound. The two are complementary. Evidence-aware confidence makes the contract enforceable, and contract enforcement converts evidence-awareness into a bound.

\subsubsection{Observability Sensitivity Diagnostic}

We measure how confidence responds to evidence reduction using a crude observability proxy. Define $\hat{\zeta} = 1$ for full evidence (18 frames) and $\hat{\zeta} = 0$ for degraded evidence (6 frames). The key diagnostic is whether the confidence distribution contracts when observability decreases:
\[
\Pr(p \geq 0.9 \mid \hat{\zeta} = 0) \quad \text{vs} \quad \Pr(p \geq 0.9 \mid \hat{\zeta} = 1)
\]

\begin{table}[h]
\centering
\small
\begin{tabular}{lcc}
\toprule
Metric & 18 frames ($\hat{\zeta}=1$) & 6 frames ($\hat{\zeta}=0$) \\
\midrule
$\Pr(\text{conf} \geq 0.9)$ & 64.1\% (189/295) & 54.8\% (160/292) \\
$\Pr(\text{wrong} \mid \text{conf} \geq 0.9)$ & 9.0\% & 8.8\% \\
Mean confidence & 0.832 & 0.804 \\
Quartiles (Q25/Q50/Q75) & 0.70 / 0.90 / 0.90 & 0.70 / 0.90 / 0.90 \\
IQR & 0.20 & 0.20 \\
\bottomrule
\end{tabular}
\caption{\small Observability sensitivity diagnostic on matched question instances ($n=295$ at 18 frames, $n=292$ at 6 frames). Despite 3$\times$ evidence reduction, the confidence distribution barely moves: quartiles are identical, IQR is identical, and median remains 0.90. Confidence does not contract commensurate with evidence loss.}
\label{tab:warrant}
\end{table}

Confidence does not contract commensurate with evidence loss. Despite reducing frames from 18 to 6, median confidence remains identical at 0.900. The high-confidence rate drops modestly (64.1\% to 54.8\%), but the error rate among high-confidence predictions is nearly identical (9.0\% vs 8.8\%). Confidence is not evidence-conditioned. The model maintains high confidence despite reduced observability, and the modest selectivity increase (coverage drops 14\% at fixed $\varepsilon$) is not proportional to the 67\% evidence reduction.

\subsection{Limitations}

Several limitations apply. We evaluate a single model (Gemini 2.0 Flash) on a single dataset (NExT-QA). Other VLMs may exhibit different confidence behaviors, and video domains beyond short activity clips may show different degradation patterns. The 300-item sample provides sufficient statistical power for mid-range $\varepsilon$ values, but estimates become noisy at extreme thresholds where few predictions pass the gate.

\paragraph{Evidence reduction is not semantic information reduction.}
Reducing frame count is not equivalent to proportionally reducing task-relevant semantic information. Videos can be temporally redundant, and many NExT-QA instances may remain answerable from sparse keyframes. No widely agreed-upon methodology exists for quantifying semantic information in a video relative to a question independent of a particular model. We complement uniform subsampling with Temporal Ablation (Appendix~\ref{app:halfclip}), which restricts frames to early or late video segments. This provides a stronger intervention: for questions about events in the missing segment, the evidence is semantically insufficient by construction, not merely sparse.

\paragraph{Self-reported confidence is a behavioral interface.}
Self-reported confidence is not guaranteed to correspond to any calibrated uncertainty estimate. It may reflect instruction-following behavior. Nevertheless, confidence-as-text is a realistic interface used in LLM/VLM deployments, and our main experiments characterize this interface's reliability under evidence truncation. Logprob Confidence uses logprob-derived confidence via the Vertex AI SDK for direct comparison of self-reported confidence against logit-derived scores ($p_{\max}$, margin, entropy). The logprob analysis shows that the model's token-level decision distribution is \emph{even less} sensitive to evidence degradation than self-reported confidence, confirming that overconfidence is representational rather than behavioral.

Both limitations reinforce the same conclusion. Warrant-like constraints should be defined over evidence-conditioned knowability and should not rely solely on a single confidence scalar, whether self-reported or logit-derived, without explicit conditioning on the evidence view.

These experiments do not estimate a warrant quantity $\zeta$ or enforce warrant-based bounds. We validate the need for such mechanisms, not their implementation.

\section{Conclusion}

We evaluated confidence-gated abstention for video question answering, testing both in-distribution behavior and robustness to evidence degradation. Confidence-based selective prediction provides mechanistic control over risk-coverage tradeoffs within the baseline distribution. Sweeping threshold $\varepsilon$ from 0 to 0.71 reduces risk from 23.6\% to 9.4\% at 63.7\% coverage, with well-calibrated predictions (ECE = 0.018).

This control is not epistemic. When frame count drops from 18 to 6, median confidence remains 0.9 and coverage drops only 16\% at fixed threshold despite a 67\% reduction in visual information. The confidence signal is not calibrated to information availability. Critically, logprob-derived confidence ($p_{\max}$) shows even less sensitivity to evidence degradation than self-reported confidence, which is inconsistent with the hypothesis that overconfidence is merely a behavioral artifact. The problem is fundamental to the model's representations.

A system deployed under variable evidence conditions cannot use a threshold tuned in-distribution. The threshold achieving 9\% error at 63.7\% coverage on full evidence yields 9\% error at only 53.7\% coverage when evidence degrades. Robust selective prediction requires making confidence evidence-aware, either through fine-tuning with observability proxies or architectural changes that condition confidence on input quality metrics such as frame count, temporal coverage, or motion density.

\section*{Reproducibility}

All experiments are fully reproducible. We use the NExT-QA validation split with 300 stratified items frozen in \texttt{item\_ids.json}. The model is Gemini 2.0 Flash (\texttt{gemini-2.0-flash}) configured with temperature 0 and max\_tokens 256. Evidence packets are extracted deterministically with SHA256 hashes recorded in manifest files for cryptographic verification. The prompt template (version v1) is stored in \texttt{config/prompts/v1.txt}. Complete provenance information including timestamps, API latencies, and raw model outputs is logged for every prediction.

\appendix
\section{Full Sweep Results}
\label{app:sweep}

\begin{table}[h]
\centering
\small
\begin{tabular}{ccccccc}
\toprule
$\varepsilon$ & Risk & Coverage & Abstention & Acc (cond) & ECE & n accepted \\
\midrule
0.00 & 0.236 & 0.987 & 0.013 & 0.764 & 0.067 & 296 \\
0.33 & 0.234 & 0.983 & 0.017 & 0.766 & 0.067 & 295 \\
0.54 & 0.226 & 0.973 & 0.027 & 0.774 & 0.062 & 292 \\
0.63 & 0.179 & 0.877 & 0.123 & 0.821 & 0.041 & 263 \\
0.71 & 0.094 & 0.637 & 0.363 & 0.906 & 0.018 & 191 \\
0.83 & 0.090 & 0.630 & 0.370 & 0.910 & 0.015 & 189 \\
0.92 & 0.019 & 0.173 & 0.827 & 0.981 & 0.009 & 52 \\
\bottomrule
\end{tabular}
\caption{Selected sweep results from Baseline Risk-Coverage. Full results in \texttt{sweep\_results.csv}.}
\label{tab:full-sweep}
\end{table}

\section{Per-Category Degradation}
\label{app:category}

We analyze degradation patterns across question categories (Causal: CW+CH, Temporal: TN+TC+TP, Descriptive: DO+DL+DC) at three operating points to prevent cherry-picking. Note that the overall coverage reported in Table 3 (e.g., 53.7\% at $\varepsilon=0.71$ under Sparse) is the average across these three 100-item strata; the per-category breakdown below explains the aggregate behavior.

\begin{table}[h]
\centering
\small
\begin{tabular}{lccccccc}
\toprule
Category & $n$ & Acc$_{18}$ & Acc$_6$ & $\Delta$Acc & Cov$_{18}$ & Cov$_6$ & $\Delta$Cov \\
\midrule
\multicolumn{8}{l}{\textit{$\varepsilon = 0$ (unconditional)}} \\
Causal & 100 & 80.0\% & 77.3\% & $-$2.7\% & 100\% & 97.0\% & $-$3.0\% \\
Temporal & 100 & 63.3\% & 60.4\% & $-$2.8\% & 98.0\% & 96.0\% & $-$2.0\% \\
Descriptive & 100 & 85.7\% & 79.8\% & $-$5.9\% & 98.0\% & 99.0\% & +1.0\% \\
\midrule
\multicolumn{8}{l}{\textit{$\varepsilon = 0.71$ (paper operating point)}} \\
Causal & 100 & 94.7\% & 97.5\% & +2.8\% & 57.0\% & 40.0\% & $-$17.0\% \\
Temporal & 100 & 83.0\% & 92.5\% & +9.5\% & 53.0\% & 40.0\% & $-$13.0\% \\
Descriptive & 100 & 92.6\% & 86.4\% & $-$6.2\% & 81.0\% & 81.0\% & 0.0\% \\
\bottomrule
\end{tabular}
\caption{Per-category accuracy and coverage at two operating points. At $\varepsilon=0$, Descriptive questions degrade most ($-$5.9\% accuracy). At $\varepsilon=0.71$, category behaviors diverge: Causal and Temporal show improved conditional accuracy but much lower coverage, while Descriptive maintains coverage but degrades accuracy.}
\label{tab:category}
\end{table}

The category analysis reveals heterogeneous degradation patterns. At $\varepsilon=0$ (unconditional), Descriptive questions show the largest accuracy drop ($-$5.9\%), followed by Temporal ($-$2.8\%) and Causal ($-$2.7\%). However, at $\varepsilon=0.71$, the pattern reverses for Causal and Temporal. Conditional accuracy \emph{increases} because the threshold more aggressively filters out uncertain predictions in the degraded regime (coverage drops from 57\% to 40\% for Causal). Descriptive questions maintain coverage but degrade accuracy, suggesting the model is overconfident on descriptive queries under evidence degradation.

\section{Temporal Ablation}
\label{app:halfclip}

Uniform subsampling reduces frame count but does not guarantee reduction of task-relevant semantic information---videos may be temporally redundant, with answers inferable from any subset of frames. To better approximate semantic information reduction, we design a procedural ablation that systematically removes temporal context by restricting evidence to specific video segments.

For each video, we generate two 6-frame evidence packets: \emph{early-half} (frames sampled uniformly from the first 50\% of the clip) and \emph{late-half} (frames sampled uniformly from the second 50\%). This design targets temporal and causal questions, which often require observing sequences of events spanning the full video. By restricting frames to one half, we remove evidence about events occurring in the other half---a more controlled intervention on semantic content than uniform subsampling, which may still capture key moments regardless of density.

\begin{table}[h]
\centering
\small
\begin{tabular}{lcccc}
\toprule
Condition & Coverage & Risk & Conditional Acc & Mean Confidence \\
\midrule
\multicolumn{5}{c}{\textit{At $\varepsilon = 0$ (unconditional)}} \\
Original (18 frames) & 98.7\% & 23.6\% & 76.4\% & 0.818 \\
Sparse (6 uniform) & 97.3\% & 27.4\% & 72.6\% & 0.786 \\
Early-half (0--50\%) & 97.3\% & 25.7\% & 74.3\% & 0.792 \\
Late-half (50--100\%) & 97.7\% & 25.3\% & 74.7\% & 0.800 \\
\midrule
\multicolumn{5}{c}{\textit{At $\varepsilon = 0.71$}} \\
Original (18 frames) & 63.7\% & 9.4\% & 90.6\% & 0.923 \\
Sparse (6 uniform) & 53.7\% & 9.3\% & 90.7\% & 0.926 \\
Early-half (0--50\%) & 52.3\% & 8.9\% & 91.1\% & 0.923 \\
Late-half (50--100\%) & 51.7\% & 10.3\% & 89.7\% & 0.929 \\
\bottomrule
\end{tabular}
\caption{\small Temporal Ablation results. At $\varepsilon=0$, early-half and late-half conditions perform similarly to uniform 6-frame sampling (25--26\% risk vs 27\%). At $\varepsilon=0.71$, all 6-frame conditions converge to similar coverage (51--54\%) with mean confidence remaining above 0.92. The model does not differentiate between early and late evidence despite the procedural removal of temporal context.}
\label{tab:halfclip}
\end{table}

Temporal Ablation conditions behave nearly identically to uniform 6-frame subsampling. At $\varepsilon=0$, early-half (25.7\% risk) and late-half (25.3\% risk) perform comparably to Sparse (27.4\% risk). At $\varepsilon=0.71$, all three 6-frame conditions converge to similar coverage (51--54\%) and mean confidence ($\sim$0.92). The model does not ``notice'' whether it is seeing the first or second half of the video. Confidence remains high regardless of which temporal segment is provided.

This finding strengthens our claim that confidence does not track evidence completeness. Unlike uniform subsampling, which may preserve key moments by chance, temporal ablation guarantees removal of one half of the clip's temporal context. For questions whose answers depend on late events (e.g., ``what does X do after Y?''), early-half evidence is semantically insufficient. Yet the model maintains similarly high confidence in both conditions. This suggests overconfidence is not merely an artifact of temporal redundancy in the dataset, but reflects a fundamental insensitivity to evidence availability.

\section{Formal Metric Definitions}
\label{app:metrics}

This appendix provides complete mathematical definitions for all evaluation metrics used in this work.

\subsection{Selective Prediction Metrics}

Let $P$ denote the full set of $n$ predictions, and let $A_\varepsilon \subseteq P$ be the subset of predictions \emph{accepted} at confidence threshold $\varepsilon$---i.e., predictions where the model's confidence $c_i \geq \varepsilon$.

\paragraph{Coverage.} The fraction of inputs for which the model provides an answer:
\begin{equation}
\text{Coverage}(\varepsilon) = \frac{|A_\varepsilon|}{|P|}
\end{equation}

\paragraph{Risk.} The error rate among accepted predictions:
\begin{equation}
\text{Risk}(\varepsilon) = \frac{1}{|A_\varepsilon|} \sum_{i \in A_\varepsilon} \mathbf{1}[\hat{y}_i \neq y_i]
\end{equation}
where $\hat{y}_i$ is the predicted answer and $y_i$ is the ground truth.

\paragraph{Conditional Accuracy.} The complement of risk:
\begin{equation}
\text{Acc}_{\text{cond}}(\varepsilon) = 1 - \text{Risk}(\varepsilon) = \frac{1}{|A_\varepsilon|} \sum_{i \in A_\varepsilon} \mathbf{1}[\hat{y}_i = y_i]
\end{equation}

\subsection{Calibration Metrics}

\paragraph{Expected Calibration Error (ECE).} Predictions are partitioned into $B$ equal-width bins by confidence. Let $B_b$ denote the set of predictions in bin $b$. ECE measures the weighted average gap between confidence and accuracy:
\begin{equation}
\text{ECE} = \sum_{b=1}^{B} \frac{|B_b|}{n} \left| \text{acc}(B_b) - \text{conf}(B_b) \right|
\end{equation}
where:
\begin{align}
\text{acc}(B_b) &= \frac{1}{|B_b|} \sum_{i \in B_b} \mathbf{1}[\hat{y}_i = y_i] \\
\text{conf}(B_b) &= \frac{1}{|B_b|} \sum_{i \in B_b} c_i
\end{align}
We use $B = 10$ bins throughout. A perfectly calibrated model has $\text{ECE} = 0$.

\subsection{Logprob-Derived Confidence Metrics}

Given raw log probabilities $\{\ell_A, \ell_B, \ell_C, \ell_D, \ell_E\}$ from the model's token distribution over the five answer options, we first normalize via softmax:
\begin{equation}
p_i = \frac{\exp(\ell_i)}{\sum_{j \in \{A,B,C,D,E\}} \exp(\ell_j)}
\end{equation}

From this distribution $\mathbf{p} = (p_A, p_B, p_C, p_D, p_E)$, we derive three confidence metrics:

\paragraph{Maximum Probability ($p_{\max}$).} The probability assigned to the most likely answer:
\begin{equation}
p_{\max} = \max_{i \in \{A,B,C,D,E\}} p_i
\end{equation}
Higher $p_{\max}$ indicates the model concentrates probability mass on a single option.

\paragraph{Margin.} The gap between the top two probabilities:
\begin{equation}
\text{Margin} = p_{\max} - p_{\text{second}}
\end{equation}
where $p_{\text{second}}$ is the second-highest probability. Larger margins indicate more decisive predictions.

\paragraph{Normalized Entropy.} Uncertainty measured as the ratio of actual entropy to maximum entropy:
\begin{equation}
\text{Entropy}_{\text{norm}} = \frac{H(\mathbf{p})}{H_{\max}} = \frac{-\sum_{i} p_i \log p_i}{\log 5}
\end{equation}
where $H_{\max} = \log 5$ is the entropy of a uniform distribution over 5 options. Values near 0 indicate high confidence (concentrated distribution); values near 1 indicate high uncertainty (uniform distribution).

\section{Prompt Templates}
\label{app:prompt}

We use two different prompts for self-reported confidence (JSON output) and logprob-derived confidence (letter-only output).

\subsection{Self-Reported Confidence Prompt}

\begin{tcolorbox}[
  colback=gray!5,
  colframe=gray!50,
  fonttitle=\bfseries\small,
  title=JSON Prompt for Self-Reported Confidence (v1),
  boxrule=0.5pt,
  arc=2pt,
  left=6pt,
  right=6pt,
  top=4pt,
  bottom=4pt
]
\small\ttfamily
You are answering a multiple-choice video question using ONLY the provided frames.

\vspace{0.5em}
\normalfont\small\textbf{Rules:}
\begin{itemize}[leftmargin=1.5em, itemsep=2pt, topsep=2pt]
  \item Use only visual evidence from the frames. Do not assume details not visible.
  \item Do not explain your reasoning. Do not write prose.
  \item If the frames do not contain enough evidence to choose confidently, abstain.
  \item If you do not abstain, you must select exactly one option A--E.
  \item Your confidence must be a number in $[0,1]$ reflecting probability of correctness.
\end{itemize}

\vspace{0.5em}
\normalfont\small\textbf{Output Format:} JSON only with this exact schema (no extra keys):

\vspace{0.3em}
\begin{lstlisting}[
  basicstyle=\ttfamily\small,
  backgroundcolor=\color{white},
  frame=none,
  xleftmargin=1em
]
{
  "choice": "A" | "B" | "C" | "D" | "E" | null,
  "confidence": <number>,
  "abstain": <boolean>,
  "evidence_span": [start_idx, end_idx] | null
}
\end{lstlisting}
\end{tcolorbox}

\subsection{Logprob Extraction Prompt}

\begin{tcolorbox}[
  colback=gray!5,
  colframe=gray!50,
  fonttitle=\bfseries\small,
  title=Letter-Only Prompt for Logprob Extraction,
  boxrule=0.5pt,
  arc=2pt,
  left=6pt,
  right=6pt,
  top=4pt,
  bottom=4pt
]
\small\ttfamily
You are answering a multiple choice question about a video. The video frames are provided in chronological order.

\vspace{0.5em}
Based on the video frames, answer the following question by selecting exactly ONE option (A, B, C, D, or E).

\vspace{0.5em}
Question: \{QUESTION\_TEXT\}

\vspace{0.5em}
Options:\\
A) \{OPT\_A\}\\
B) \{OPT\_B\}\\
C) \{OPT\_C\}\\
D) \{OPT\_D\}\\
E) \{OPT\_E\}

\vspace{0.5em}
\textbf{Respond with ONLY a single letter (A, B, C, D, or E) and nothing else.}
\end{tcolorbox}

\vspace{0.5em}
\noindent The letter-only prompt is used with \texttt{response\_logprobs=True} to extract token-level probabilities over answer options. This simpler output format avoids JSON parsing and enables direct access to the model's softmax distribution over the five answer tokens.

\subsection{Logprob-to-Probability Computation}

\begin{figure}[h]
    \centering
    \includegraphics[width=0.75\textwidth]{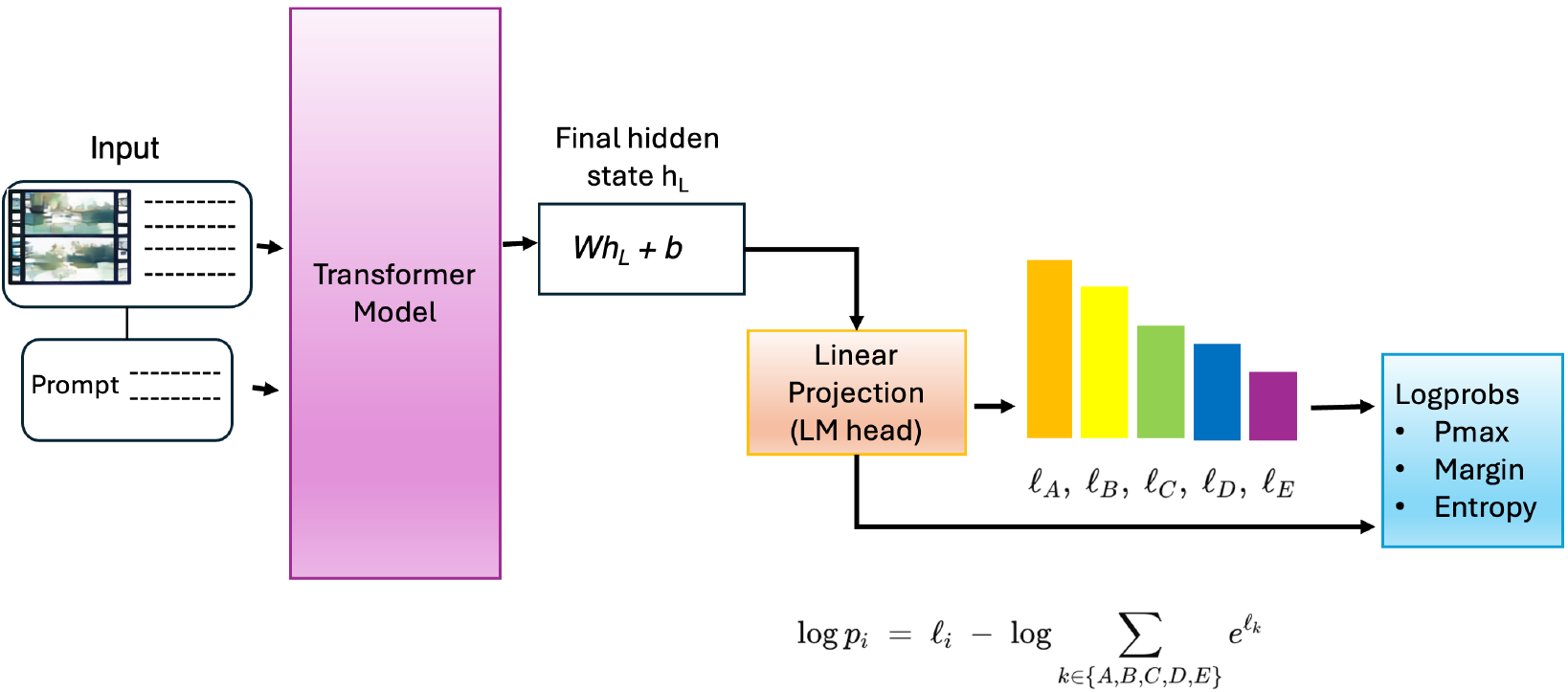}
    \caption{Logprob-to-probability computation pipeline. Video frames and prompt are processed by the transformer model. The final hidden state is projected via the LM head to produce logits over the vocabulary. We extract logprobs for tokens A--E and renormalize via softmax to obtain a probability distribution over answer options, from which we derive $p_{\max}$, margin, and entropy.}
    \label{fig:logprob-computation}
\end{figure}

Figure~\ref{fig:logprob-computation} illustrates the logprob extraction pipeline. Raw log probabilities $\ell_i$ from the model's token distribution are normalized via softmax over the five answer options A--E. This renormalization is necessary because the model's full vocabulary distribution includes tokens beyond the answer options; we restrict to the answer space to compute a proper probability distribution. The resulting $p_i$ values sum to 1 and represent the model's relative preference among answer choices.

\bibliographystyle{unsrtnat}
\bibliography{references}

\end{document}